\documentclass[]{spie}  %>>> use for US letter paper
\usepackage{amsmath,amsfonts,amssymb}
\usepackage{graphicx}
\usepackage[colorlinks=true, allcolors=blue]{hyperref}
\usepackage{makecell}
\usepackage{multirow}
\usepackage{tabularx}
\usepackage{graphicx}
\usepackage{subcaption}

\title{Automated Estimation of Anatomical Risk Metrics for Endoscopic Sinus Surgery Using Deep Learning}

\author[a,b]{Konrad Reuter}
\author[a]{Lennart Thaysen}
\author[b]{Bilkay Doruk}
\author[a]{Sarah Latus}
\author[b]{Brigitte Holst}
\author[b]{Benjamin Becker}
\author[b]{Dennis Eggert}
\author[b]{Christian Betz}
\author[b]{Anna-Sophie Hoffmann*}
\author[a]{Alexander Schlaefer*}
\affil[a]{Hamburg University of Technology, Hamburg, Germany}
\affil[b]{University Medical Center Hamburg-Eppendorf, Hamburg, Germany}

\authorinfo{Further author information: (Send correspondence to K.Reuter)\\ K. Reuter: E-mail: konrad.reuter@tuhh.de, Telephone: +49 40 30601 2858\\ *These authors contributed equally to this work.}

\begin{document} 
\maketitle

\begin{abstract}
Endoscopic sinus surgery requires careful preoperative assessment of the skull base anatomy to minimize risks such as cerebrospinal fluid leakage. Anatomical risk scores like the Keros, Gera and Thailand-Malaysia-Singapore score offer a standardized approach but require time-consuming manual measurements on coronal CT or CBCT scans. We propose an automated deep learning pipeline that estimates these risk scores by localizing key anatomical landmarks via heatmap regression. We compare a direct approach to a specialized global-to-local learning strategy and find mean absolute errors on the relevant anatomical measurements of $0.506$ mm for the Keros, $4.516$\textdegree\ for the Gera and $0.802$ mm / $0.777$ mm for the TMS classification.
\end{abstract}

% Include a list of keywords after the abstract 

\section{INTRODUCTION}
\label{sec:intro}  % \label{} allows reference to this section

\begin{figure}[ht]
  \centering
  \begin{minipage}[b]{0.25\textwidth}
    \centering
    \includegraphics[width=\linewidth]{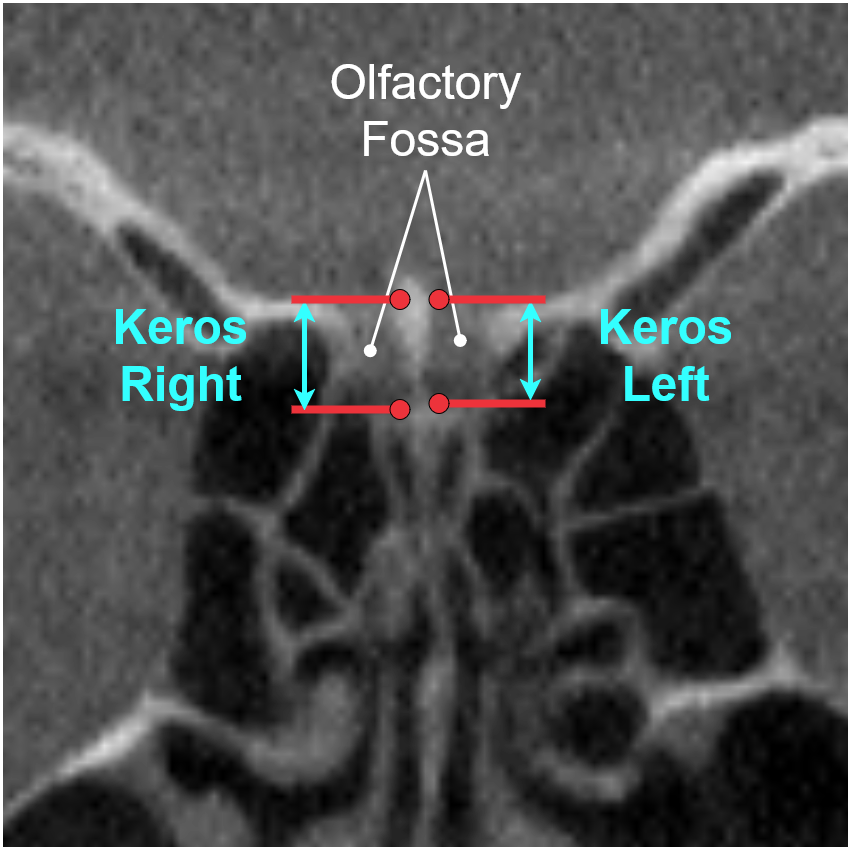}
  \end{minipage}
  \hspace{0.01\textwidth}
  \begin{minipage}[b]{0.25\textwidth}
    \centering
    \includegraphics[width=\linewidth]{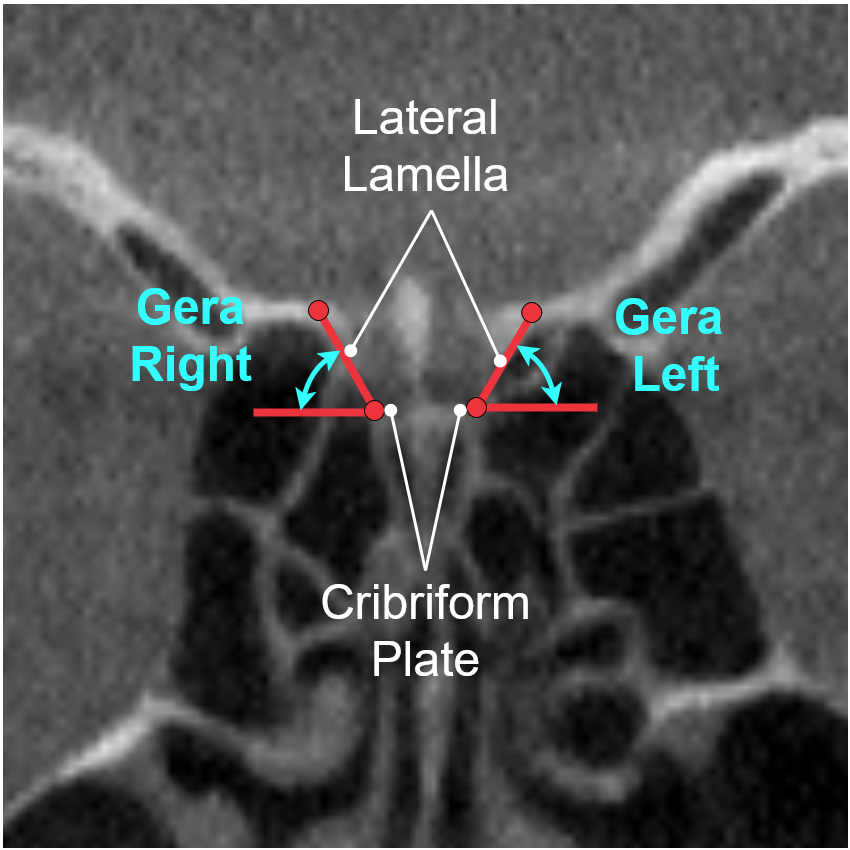}
  \end{minipage}
  \hspace{0.01\textwidth}
  \begin{minipage}[b]{0.25\textwidth}
    \centering
    \includegraphics[width=\linewidth]{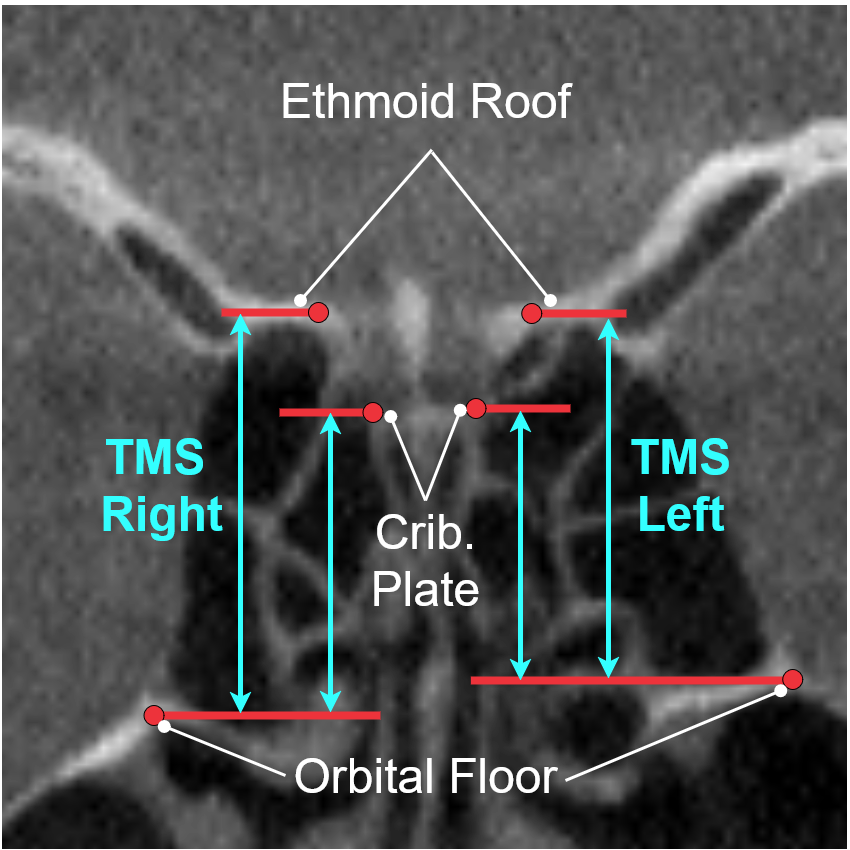}
  \end{minipage}
  \vspace{0.1cm}
  \caption{Visual Representation of the Keros (left), Gera (middle) and TMS (right) Score. The red dots represent the points to be predicted by our models.}
    \label{fig:classifications}

\end{figure}

Endoscopic sinus surgery (ESS) is a widely used, minimally invasive technique for accessing and treating diseases of the paranasal sinuses and adjacent structures. It is primarily indicated in conditions such as chronic rhinosinusitis, particularly when refractory to medical therapy \cite{ESS_chronic}. Additional indications include management of nasal polyps, certain tumors, and orbital or intracranial complications \cite{ESS_review}. Due to the close proximity of the sinuses to critical anatomical structures, ESS comes with inherent risks. One major concern is potential damage to the skull base, which may result in cerebrospinal fluid leakage \cite{ESS_complications}. To minimize such risks, a preoperative assessment of the patients anatomy is essential. Over the years, different metrics have been developed to quantify anatomical risk and surgical complexity, typically derived from the coronal cross-section of the anterior skull in CBCT or CT scans. A well-known early method is the Keros classification \cite{keros}. It differentiates between three risk classes based on the depth of the olfactory fossa, which is measured as the vertical distance between the fovea ethmoidalis and the cribriform plate. However, the Keros Score alone has been found insufficient to fully describe the anatomy of the skull base and potential risks \cite{Keros_limitations}. In response, Gera et al. proposed to analyze the slope of the skull base in the coronal plane and classify based on the angle defined by the lateral lamella and the horizontal plane passing through the cribriform plate \cite{gera}. Recognizing that the Keros and Gera classification are impractical for an intraoperative assessment and may not consider anatomical differences across populations, the Thailand-Malaysia-Singapore (TMS) Score was proposed \cite{tms}. Using the orbital floor as a reference point, this method uses the lateral distances to the ethmoid roof as well as the cribriform plate for classification. Importantly, the authors of the TMS classification consider it as a complement to, rather than a replacement for the Keros and Gera scores. We summarize the three classification methods in Table \ref{tab:classifications} and visualize them in Figure \ref{fig:classifications}. Given the increasing complexity and time required to apply these classifications in combination, an automated solution could save valuable time. In addition, it could assist less experienced surgeons during preoperative planning by highlighting potentially challenging cases that might require referral to specialized centers, while straightforward cases could be operated safely in the routine clinical settings. To this end, we present the, to the best of our knowledge, first automated approach for evaluating the Keros, Gera, and TMS scores using deep learning. We assess well-established models to identify relevant anatomical structures in coronal CBCT slices, aiming to streamline risk classification and improve surgical planning. Our global-to-local learning strategy first localizes relevant regions before performing detailed predicions on focused local patches. This two-stage approach not only improves performance but also stabilizes results across all models compared to direct, single-stage methods.

\vspace{-0.1cm}

\vspace{-0.1cm}
\begin{table}[ht]
\centering
\caption{Comparison of Keros, Gera, and TMS Classifications}
\label{tab:classifications}
\begin{tabular}{|c|c|c|c|c|}
\hline
\textbf{Classification} & \textbf{Anatomical Feature} & \textbf{Class I} & \textbf{Class II} & \textbf{Class III} \\
\hline
\textbf{Keros} & \makecell{Depth of the olfactory fossa}  & \textless 4 mm & 4-8 mm & \textgreater 8 mm \\
\hline
\textbf{Gera} & \makecell{Angle between lateral lamella and \\ horizontal plane through cribriform plate} & \textgreater80\textdegree & 45–80\textdegree &  \textless45\textdegree \\
\hline
\textbf{TMS} & \makecell{1. Lateral distance orbital floor to cribriform plate \\ 2. Lateral distance orbital floor to ethmoid roof} & \makecell{both \\ \textgreater 10 mm} & \makecell{one \\ \textless 10 mm} & \makecell{both \\ \textless 10mm} \\
\hline
\end{tabular}
\end{table}
\vspace{-0.1cm}
\section{MATERIAL AND METHODS}
\vspace{-0.1cm}
\subsection{Dataset}

The dataset was acquired at the University Medical Center Hamburg-Eppendorf between 2021 and 2023. It consists of 691 fully annotated coronal slices from CBCT scans of 640 unique patients. Most slices have a pixel spacing of 0.45 mm, only 12 cases were acquired at a higher resolution of 0.3 mm per pixel. Image dimensions range from 223 and 668 pixels in height and 245 and 668 pixels in width, with height always equal or less than width. Labeling was initially performed by a medical student trained by a senior consultant and subsequently reviewed and verified by an experienced radiologist. The annotations consist of ten 2D points per slice marking the relevant anatomical landmarks. The distribution of Keros, Gera and TMS classes are presented in Table \ref{tab:data_stats}. We adopt a heatmap regression approach to predict the anatomical landmarks. Therefore, each annotated point is converted in to a heatmap using the following equation.
\vspace{-0.1cm}
\begin{equation}
    H(x, y) = exp\left(-\left(\frac{(x-x_0)^2}{2 \sigma_x^2}+\frac{(y-y_0)^2}{2\sigma_y^2}\right)\right)
\end{equation}

\vspace{-0.1cm}

where $\sigma_x$ and $\sigma_y$ define the shape of the Gaussian probability distribution around the target point $(x, y)$. We restore the point coordinates from a heatmap by calculating the center of mass in a window of $13 \times 13$ pixels around the maximum value.

\begin{table}[hb]
    \centering
            \caption{Distribution of Keros, Gera and TMS classes in our dataset.}
    \begin{tabular}{|c|c|c|c|}

         %Class & Keros Left & Keros Right & Gera Left & Gera Right & TMS Left & TMS Right \\
         \hline
         \textbf{Class} & \textbf{Keros} & \textbf{Gera} & \textbf{TMS} \\
         \hline 
         \textbf{I} & 269 (19.46 \%) & 43 (3.11 \%) & 1193 (86.32 \%)\\
         \textbf{II} & 970 (70.19 \%) & 1276 (92.33 \%) & 181 (13.10 \%) \\
         \textbf{III} & 143 (10.35 \%) & 63 (4.55 \%)  & 8 (0.58 \%) \\
         \hline
         %1 & 139 & 130 & 19 & 24 & 608 & 585 \\
         %2 & 480 & 490 & 637 & 639 & 81 & 100 \\
         %3 & 72 & 71 & 35 & 28 & 2 & 6 \\
    \end{tabular}
    \label{tab:data_stats}
\end{table}

\subsection{Deep Learning Approach}

\begin{figure}
    \centering
    \includegraphics[width=0.8\linewidth]{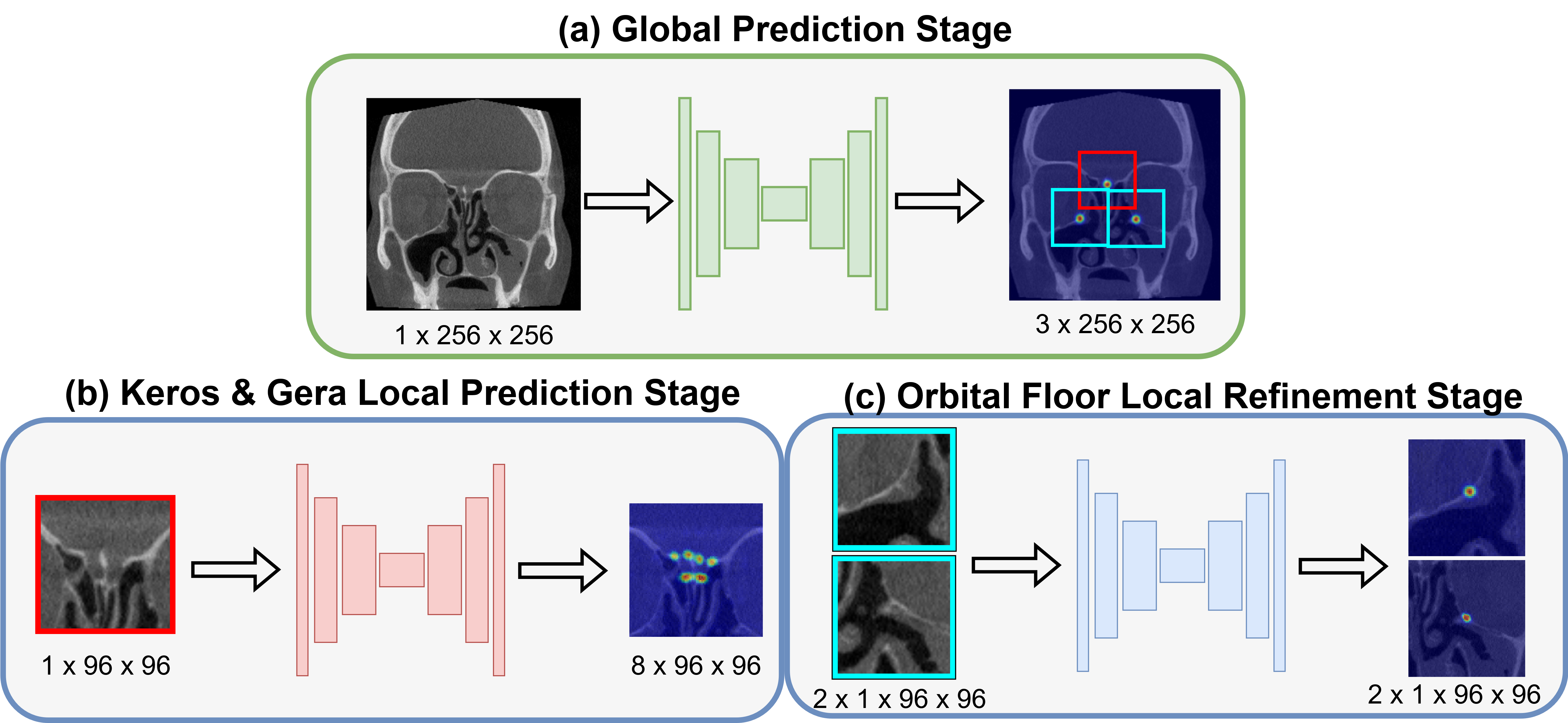}
    \vspace{0.1cm}
    \caption{Method overview. A global model (a) is used to generate position estimates of the center point of the Keros and Gera scores (red box), as well as the left and right orbital floor (blue boxes). Subsequently, two local models operate on patches extracted around the global position estimates. The first local model (b) estimates the anatomical landmarks for Keros and Gera. The second local model (c) refines the orbital floor position estimates.}
    \label{fig:method}
\end{figure}

The relevant target regions constitute only a small portion of the slices, while the majority can be considered background. However, the high anatomical variance makes simple cropping unsuitable. To address this, we adopt a global-to-local (G2L) learning strategy. We define three global reference points: the center point of the Keros and Gera landmarks, the left orbital floor, and the right orbital floor. A global model first predicts these coarse locations (Figure \ref{fig:method} a). Subsequently, two local models are applied to high-resolution image patches cropped around the predicted points. The first local model is used to estimate the anatomical landmarks for Keros and Gera (Figure \ref{fig:method} b). The second local model is used refine the position estimates of the orbital floor landmarks, treating the left and right sides as separate samples (Figure \ref{fig:method} c). 

For the global model, the images are first cropped into squares and afterwards resized to $256 \times 256$ pixels. We apply data augmentation, including random horizontal flipping, considering the symmetry of the head, as well as random rotations between -5 \textdegree and +5 \textdegree, representing slight misalignments during data acquisition. At the local stage, patches of size $96 \times 96$ pixels centered around the global positions are extracted from the original images. To improve robustness against inaccuracies in the global predictions, we add random jitter to the global positions, shifting them by up to $\pm 20$ pixels in horizontal and vertical directions before extracting the local patches. 

We evaluate four different model architectures: UNet \cite{unet} as a standard baseline, ResUNet\cite{resunet}, a 2D version of MedNeXt \cite{mednext}, and a 2D version of SwinUNETR-V2 \cite{swinunetrv2} as a transformer based alternative. All models are configured to have a comparable amount of parameters. We compare our G2L method to a direct approach, where we train the models to predict the target locations on a global level. Each model is trained using a standard configuration of hyperparameters: a batch size of 16, AdamW optimizer with a learning rate of $1\cdot10^{-4}$ and MSE loss. $\sigma_x$ and $\sigma_y$ are set to 2 for the local models and the direct approach, and to 4 for the global models. Early stopping with a patience of 20 epochs is employed to prevent overfitting. The dataset is split into 6 groups with equal number of patients. One group is used for testing, while the remaining are utilized for five-fold cross-validation.

We evaluate the global models using the Euclidean distance between predicted and groundtruth positions, reported as the mean absolute error (MAE) and the maximum error (MAXE). For the direct and G2L approach we present the MAEs for the relevant anatomical distances (Keros, TMS) or angles (Gera).

\section{RESULTS}

All tables report mean results over 5-fold cross-validation. Table \ref{tab:global_metrics} summarizes the global model performance. UNet achieves the lowest average error, whereas ResUNet yields the lower maximum error. Since minimizing the maximum error is more critical at the global stage, we selected ResUNet to generate the G2L results in Table \ref{tab:model_metrics_clean}. For Keros and Gera, the G2L approach consistently reduces MAE across all models, with the largest gains for SwinUNETR and MedNeXt, bringing their performance close to that of UNet and ResUNet. For TMS, all models improve except UNet. Class based metrics (Table \ref{tab:classbased}), reported with ResUNet for both global and local stages, reveal only a slight bias toward overrepresented classes in Keros (Class II) and TMS (Class I), but a clear bias in Gera (Class II).

\noindent

\begin{minipage}[t]{0.36\textwidth}
\centering
\captionof{table}{Global model performance}
\begin{tabular}{|l|c|c|}
\hline
\textbf{Model} & \textbf{\makecell{MAE \\ {[mm]}}} & \textbf{\makecell{MAXE \\ {[mm]}}} \\
\hline 
UNet        & \textbf{1.023} & 9.95 \\
ResUNet & 1.161 & \textbf{6.94} \\
SwinUNETR-V2 & 1.751 & 15.26 \\
MedNeXt     & 1.927 & 18.41 \\
\hline
\end{tabular}
\label{tab:global_metrics}

\end{minipage}
\hfill
\begin{minipage}[t]{0.62\textwidth}
\centering
\captionof{table}{Performance metrics of the direct and G2L approaches.}
\begin{tabular}{|c|l|c|c|c|c|}
\hline
&\textbf{Model}  & \textbf{\makecell{Keros \\ MAE \\ {[mm]}}} & \textbf{\makecell{Gera \\ MAE\\ {[\textdegree]}}} & \textbf{\makecell{TMS1 \\ MAE\\ {[mm]}}} & \textbf{\makecell{TMS2 \\ MAE\\ {[mm]}}} \\
\hline
\multirow{4}{*}{\rotatebox[origin=c]{90}{\textbf{Direct}}}&
UNet       & 0.602 & 5.040 & 0.858 & \textbf{0.777} \\
&ResUNet & 0.669 & 5.319 & 0.966 & 0.898 \\
&SwinUNETR-V2  & 0.996 & 7.678 & 1.641 & 1.636  \\
&MedNeXt    & 1.156 & 7.151 & 1.526 & 1.387 \\
\hline
\multirow{4}{*}{\rotatebox[origin=c]{90}{\textbf{G2L}}}& 
UNet       & 0.530 & \textbf{4.516} & 0.861 & 0.831 \\
&ResUNet & 0.513 & 4.717 & \textbf{0.802} & 0.793 \\
&SwinUNETR-V2 & \textbf{0.506} & 5.075 & 0.852 & 0.780 \\
&MedNeXt   & 0.556 & 4.906 & 0.912 & 0.850 \\
\hline
\end{tabular}
\label{tab:model_metrics_clean}
\end{minipage}

\begin{table}[ht]
    \centering
        \caption{Class based precision, recall and accuracy for the G2L approach using ResUNet for the global and local stages.}
    \begin{tabular}{|c|c|c|c|c|c|c|c|c|c|}
    \hline
        &\multicolumn{3}{c|}{\textbf{Keros}} & \multicolumn{3}{c|}{\textbf{Gera}} & \multicolumn{3}{c|}{\textbf{TMS}}\\
          & I & II & III & I & II & III & I & II & III \\
         \hline
         \textbf{Precision} & 0.795 & 0.944 & 0.865 & 0.640 & 0.964 & 0.393 & 0.982 & 0.814 & 0.933 \\
         \hline
         \textbf{Recall} & 0.845 & 0.954 & 0.700 & 0.300 & 0.981 & 0.450 & 0.970 & 0.887 & 0.800 \\
          \hline
          \textbf{Accuracy} & \multicolumn{3}{c|}{0.916} & \multicolumn{3}{c|}{0.947} & \multicolumn{3}{c|}{0.956} \\
         \hline
    \end{tabular}
    \label{tab:classbased}
\end{table}

\iffalse
\begin{figure}[ht]
    \centering
    \includegraphics[width=.7\linewidth]{figures/failures.png}
    \caption{Cases with misplaced anatomical landmarks. Results generated using the G2L UNet method.}
    \label{fig:fails}
\end{figure}
\fi

\section{DISCUSSION AND CONCLUSION}

This work presents a first approach for automated prediction of the Keros, Gera and TMS classes using deep learning. We investigate a global-to-local learning strategy that involves a two-stage prediction of landmark positions, first at a global level and subsequently within local patches. We vary the model architectures in both stages and compare the G2L performance to a direct learning approach. Using ResUNet for the global and local stages yields promising results in identifying relevant anatomical landmarks with accuracies over 90 \% for all three classification methods. 
The proposed G2L strategy improved performance across all models and reduced performance variability between architectures, suggesting that it effectively helps models focus on the relevant regions. However, class based analysis reveals that the approach performs best on the most frequent classes. This effect is partly due to the joint prediction of multiple landmarks, which encourages the model to learn typical inter-landmark distances and angles. Independent prediction may help improve performance on underrepresented classes. Future work could extend this approach to 3D to automate slice selection and incorporate additional anatomical context for more robustness. Furthermore, it is important to evaluate the generalizability of the proposed approach on data from additional medical centers.  

\acknowledgments % equivalent to \section*{ACKNOWLEDGMENTS}       
 
This research was co-funded by the MARLOC project (DFG, grant SCHL 1844-10-1) and by the European Union under Horizon Europe programme grant agreement No. 101059903; and by the European Union funds for the period 2021-2027.  

% References
\bibliography{report} % bibliography data in report.bib
\bibliographystyle{spiebib} % makes bibtex use spiebib.bst

\end{document}